\pdfoutput=1
\documentclass[11pt]{article}

\usepackage{EMNLP2023}

\usepackage{times}
\usepackage{latexsym}
\usepackage{amsmath}
\usepackage{graphicx}
\usepackage{multirow}
\usepackage{multicol}
\usepackage{arydshln}
\usepackage[hang,flushmargin]{footmisc}
\usepackage{algorithm}
\usepackage{algpseudocode}
\usepackage{booktabs}
\usepackage{courier}
\usepackage{array}
\usepackage{arydshln}
\usepackage{soul}
\usepackage{subcaption}
\usepackage{url}

\usepackage{textcase}

\newcolumntype{C}[1]{>{\centering\arraybackslash}p{#1}}

\usepackage[T1]{fontenc}
\usepackage[utf8]{inputenc}

\usepackage{microtype}

\usepackage{inconsolata}

\title{\method: Data-efficient Multilingual Learning}

\author{Simran Khanuja \quad Srinivas Gowriraj \quad Lucio Dery \quad Graham Neubig \\ \\ \vspace{1mm} Carnegie Mellon University \\ \texttt{\{skhanuja,sgowrira,ldery,gneubig\}@cs.cmu.edu}}
\begin{document}
\newcommand{\refalg}[1]{Algorithm \ref{#1}}
\newcommand{\refeqn}[1]{Equation \ref{#1}}
\newcommand{\reffig}[1]{Figure \ref{#1}}
\newcommand{\reftbl}[1]{Table \ref{#1}}
\newcommand{\refsec}[1]{Section \ref{#1}}
\newcommand{\refapp}[1]{Appendix \ref{#1}}
\definecolor{coralpink}{rgb}{0.97, 0.51, 0.47}

\newcommand{\todo}[1]{\textcolor{red}{[[ #1 ]]}\typeout{#1}}

\newcommand{\m}[1]{\mathcal{#1}}
\newcommand{\bmm}[1]{\bm{\mathcal{#1}}}
\newcommand{\real}[1]{\mathbb{R}^{#1}}
\newcommand{\method}{\textsc{DeMuX}}

\newtheorem{theorem}{Theorem}[section]
\newtheorem{claim}[theorem]{Claim}

\newcommand{\argmax}{arg\,max}
\newcommand\norm[1]{\left\lVert#1\right\rVert}

\newcommand{\note}[1]{\textcolor{blue}{#1}}

\newcommand*{\Scale}[2][4]{\scalebox{#1}{$#2$}}%
\newcommand*{\Resize}[2]{\resizebox{#1}{!}{$#2$}}%

\newcommand{\gn}[1]{\textcolor{olive} {[\textsc{gn}: #1]}}
\newcommand{\simi}[1]{\textcolor{cyan} {[\textsc{sk}: #1]}}
\newcommand{\ld}[1]{\textcolor{orange} {[\textsc{ld}: #1]}}
\newcommand{\deprecated}[1]{\textcolor{coralpink}{#1}}

% \newcolumntype{L}[1]{>{\raggedright\arraybackslash}p{#1}}
% \newcolumntype{C}[1]{>{\centering\arraybackslash}p{#1}}
% \newcolumntype{R}[1]{>{\raggedleft\arraybackslash}p{#1}}
% \newcolumntype{Y}{>{\centering\arraybackslash}X}
\newcolumntype{L}{>{\centering\arraybackslash}m{3cm}}

\maketitle

\begin{abstract}
We consider the task of optimally fine-tuning pre-trained multilingual models, given small amounts of unlabelled target data and an annotation budget. In this paper, we introduce \method, a framework that prescribes the exact data-points to label from vast amounts of unlabelled multilingual data, having unknown degrees of overlap with the target set. Unlike most prior works, our end-to-end framework is language-agnostic, accounts for model representations, and supports multilingual target configurations. Our active learning strategies rely upon distance and uncertainty measures to select task-specific neighbors that are most informative to label, given a model. \method \space outperforms strong baselines in 84\% of the test cases, in the zero-shot setting of disjoint source and target language sets (including multilingual target pools), across three models and four tasks. Notably, in low-budget settings (5-100 examples), we observe gains of up to 8-11 F1 points for token-level tasks, and 2-5 F1 for complex tasks. Our code is released here\footnote{\url{https://github.com/simran-khanuja/demux}}.

% \ld{Updated comment : Yes ! I would say maybe give a sneak preview of how the method works : "a framework that prescribes the exact data-points to label from vast amounts of unlabelled multilingual data  .... " by considering uncertainty based measures directly in the embedding space of the MT model.
% }\simi{hmm I tried but I'm afraid the abstract becomes too long and too much to unpack, I'm not sure wdyt?}

% Past approaches have focused on identifying the best source languages for transfer or emphasized few-shot annotation in the target language. In this paper, we introduce \todo{\method}, a framework that prescribes the exact data-points to label from vast amounts of unlabelled multilingual data. \st{, that may or may not overlap with the target set.} Unlike most prior works, our end-to-end framework is language-agnostic, accounts for model representations, and supports multilingual target configurations.

% \st{Past approaches have focused on identifying the best source languages for transfer by relying on linguistic feature information, but do not address what data to label.} \simi{LD-sorry your comment got deleted, but is this rephrasing better?}

% Notably, in low-budget settings (5-100 examples), we observe gains of upto 8-11 F1 points for token-level tasks, and 2-5 F1 for complex tasks like NLI and QA. For some tasks, our strategies even surpass the gold standard of fine-tuning in target languages, while being zero-shot

 % \gn{The results described below seem a bit extensive for an abstract. We could make them a bit more concise.} 
\end{abstract}

\section{Introduction}
\label{intro}

\noindent Picture this: Company \textbf{Y}, a healthcare technology firm in India, has recently expanded their virtual assistance services to cover remote locations in Nepal and Bhutan. Unfortunately, their custom-trained virtual assistant is struggling with the influx of new multilingual data, most of which is in Dzongkha and Tharu, but unindentifiable by non-native researchers at \textbf{Y}. How can they improve this model? Following current approaches, they first attempt to discern the languages the data belongs to, but commercial language identification systems (LangID) are incapable of this task\footnote{For instance, this is true of Google Cloud's LangID as of November 2023: \url{https://developers.google.com/ml-kit/language/identification/langid-support}}. Assuming this hurdle is crossed, Company \textbf{Y} then seeks out annotators fluent in these languages, but this also fails given crowd-sourcing platforms' lack of support for the above languages\footnote{Given MTurk's lack of regional support.}. As an alternative, they decide to use tools that identify best languages for transfer, but these either rely on linguistic feature information -- missing for Dzongkha and Tharu\footnote{e.g. in the WALS database \url{https://wals.info/languoid}}  \cite{lin2019choosing}, past model performances -- expensive to obtain \cite{srinivasan2022litmus} or don't support multilingual targets \cite{lin2019choosing,kumar2022diversity}. Based on annotator availability, they eventually choose Nepali and Tibetan as optimal transfer languages, and collect unlabelled corpora from news articles, social media and online documents. Even assuming all the preceding challenges are surmounted, a final question remains unaddressed by the traditional pipeline: \emph{how do they select the exact data points to give to annotators for best performance in their domain-specific custom model, under a fixed budget?}

 \begin{figure*}[!htbp]
    \includegraphics[width=\textwidth]{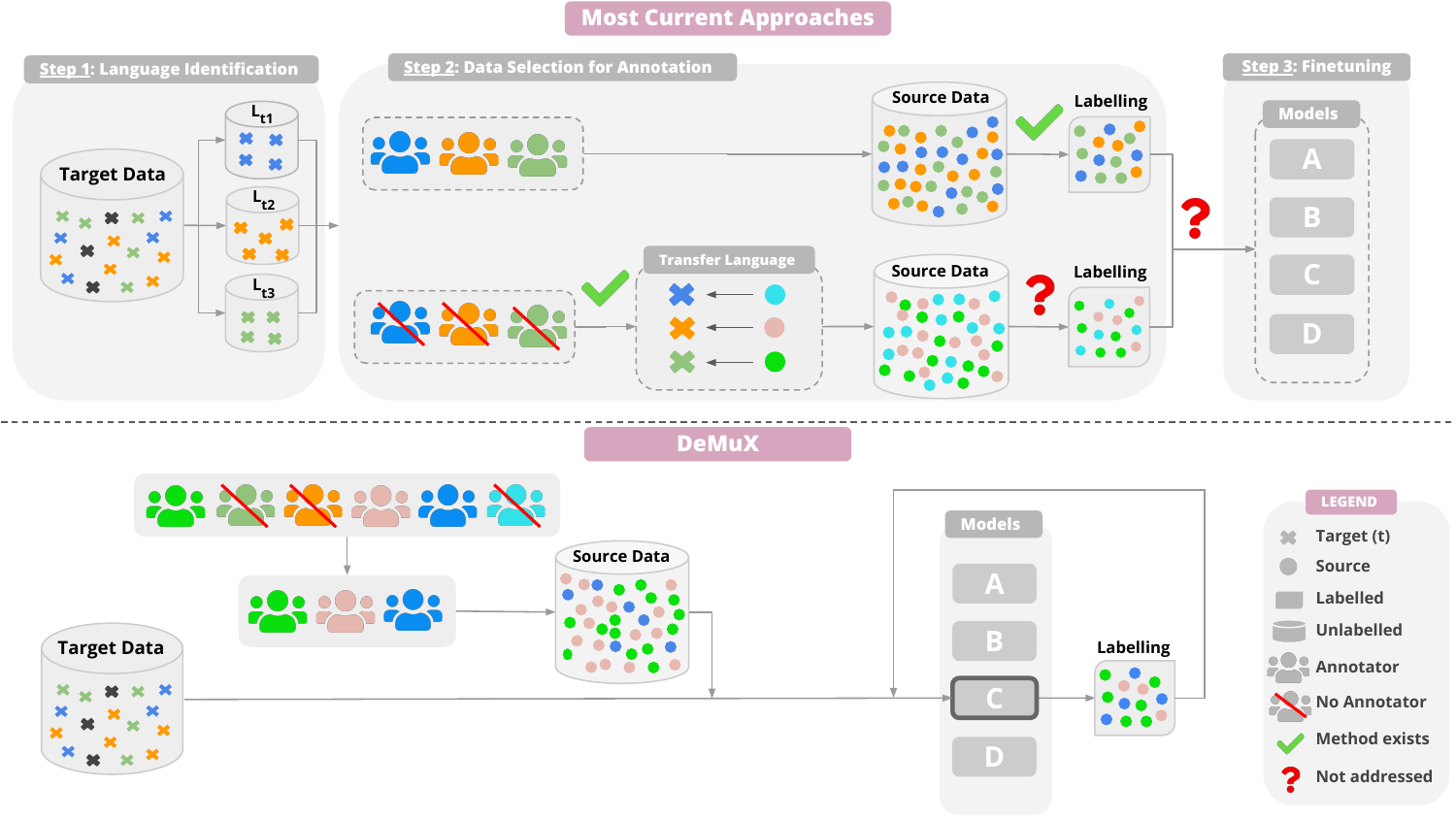}
    \caption{\emph{Top:} Today, improving the performance of a model on multilingual target data is a three-step process. First, one would identify the target languages. Next, they would either collect data to label in these languages, or closely related transfer languages, based on annotator availability. Finally, they would fine-tune a model on the labelled data. However \emph{Step 1} excludes 98\% of the world's languages, \emph{Step 2} is constrained by annotators or linguistic feature information, and \emph{Step 3} of factoring in the model to fine-tune, largely remains unaccounted for. 
    \emph{Bottom:} With our end-to-end framework \method, we prescribe the exact data to label from a vast pool of multilingual source data, that provides for best transfer to the target, for a given model.}  
    \label{fig:fig_1}
\end{figure*}

In this work, we aim to provide a solution to the above problem by introducing \method{}, an end-to-end framework that replaces the pipelined approach to multilingual data annotation (Figure \ref{fig:fig_1}). By directly selecting datapoints to annotate, \method \space bypasses several stages of the pipeline, that are barriers for most languages. The alleviation of needing to identify the target languages itself (\emph{Step 1}: Figure \ref{fig:fig_1}), implies that it can be used for noisy, unidentifiable, or code-mixed targets. 
% This remedies the cascading errors that result from using LangID systems to first identify the languages to which target data belongs. 

% \method{}'s design eliminates multiple failure points and directly selects instances to label from a large pool of unlabelled multilingual data

\method \space makes decisions at the instance level by using information about the pre-trained multilingual language model's (MultiLM's) representation space. This ensures that the data annotation process is aware of the model ultimately being utilized. Concretely, we draw from the principles of active learning (AL) \cite{cohn1996active,settles2009active} for guidance on model-aware criteria for point selection. AL aims to identify the most informative points (to a specific model) to label from a stream of unlabelled source data. Through iterations of model training, data acquisition and human annotation, the goal is to achieve satisfactory performance on a target test set, labelling only a small fraction of the data. Past works \cite{chaudhary2019little,kumar2022diversity,moniz2022efficiently} have leveraged AL in the special case where the same language(s) constitute the source and target set  (\emph{Step 2 (upper branch)}: Figure \ref{fig:fig_1}). However, none so far have considered the case of source and target languages having unknown degrees of overlap; a far more pervasive problem for real-world applications that commonly build classifiers on multi-domain data \cite{dredze2008online}. From the AL lens, this is particularly challenging since conventional strategies of choosing the most uncertain samples \cite{settles2009active}, could pick distracting examples from very dissimilar language distributions \cite{longpre2022active}. Our strategies are designed to deal with this distribution shift by leveraging small amounts of unlabelled data in target languages. 

In the rest of the paper, we first describe three AL strategies based on the principles of a) semantic similarity with the target; b) uncertainty; and c) a combination of the two, which picks uncertain points in target points' local neighborhood (\S\ref{sec:strategies}). We experiment with tasks of varying complexity, categorized based on their label structure: token-level (NER and POS), sequence-level (NLI), and question answering (QA). We test our strategies in a zero-shot setting across three MultiLMs and five target language configurations, for a budget of 10,000 examples acquired in five AL rounds (\S\ref{sec:experiments}).

We find that our strategies outperform previous baselines in most cases, including those with multilingual target sets. The extent varies, based on the budget, the task, the languages and models (\S\ref{sec:results}). Overall, we observe that the hybrid strategy performs best for token-level tasks, but picking globally uncertain points gains precedence for NLI and QA.
% \ld{Unsure if this is relevant for intro -- could remove if space is issue : }
% \st{The hybrid strategy performs best for token level tasks, but picking globally uncertain points gains precedence for sequence-level tasks and QA. Data selection also varies for the same target across tasks and models, that makes for interesting analyses of underlying representations} (\S\ref{sec:results}).
To test the applicability of \method \space in resource constrained settings, we experiment with lower budgets ranging from 5-1000 examples, acquired in a single AL round. In this setting, we observe gains of upto 8-11 F1 points for token-level tasks, and 2-5 F1 for complex tasks like NLI and QA. For NLI, our strategies surpass the gold standard of fine-tuning in target languages, while being zero-shot. 

\section{Notation}
% \gn{Maybe we can split this problem setting off as a separate section directly after the intro.}\simi{done}x
Assume that we have a set of source languages, ${L_{s}} = \{l^{1}_{s} \ldots l^{n}_{s}\}$, and a set of target languages, ${L_{t}} = \{l^{1}_{t} \ldots l^{m}_{t}\}$. ${L_{s}}$ and ${L_{t}}$ are assumed to have unknown degrees of overlap. 

Further, let us denote the corpus of unlabelled source data as $\mathcal{X}_{s} = \{x^{1}_{s} \ldots x^{N}_{s}\}$ and the unlabelled target data as $\mathcal{X}_{t} = \{x^{1}_{t} \ldots x^{M}_{t}\}$.

Our objective is to label a total budget of ${B}$ data points over ${K}$ AL rounds from the source data. The points to select in each round can then be calculated by $b = \frac{B}{K}$. Thus considering the super-set $\mathcal{S}^{b} = \{\mathbf{X} \subset \mathcal{X}_{s} ~\big|~ \lvert \mathbf{X} \rvert = b \}$ of all $b$-sized subsets of $\mathcal{X}_{s}$, our objective is to select some $\mathbf{X}^{*} \in \mathcal{S}^{b}$ according to an appropriate criterion.

\section{Annotation Strategies}
\label{sec:strategies}
Based on the broad categorizations of AL methods as defined by \citet{zhang2022survey}, we design three annotation strategies that are either \emph{representation-based}, \emph{information-based}, or \emph{hybrid}. The first picks instances that capture the diversity of the dataset; the second picks the most uncertain points which are informative to learn a robust decision boundary; and the third focuses on optimally combining both criteria.
In contrast to the standard AL setup, there are two added complexities in our framework: a) source-target domain mismatch; b) multiple distributions for each of our target languages. We therefore design our measures to select samples that are semantically similar (from the perspective of the MultiLM) to the target domain \cite{longpre2022active}.
% \gn{The following is verbose, and could be mostly removed to save space if necessary (just mention that we'll describe below).}
% A brief summary of the strategies is as follows:  
% \begin{enumerate}
%     \item \texttt{AVERAGE-DIST}: labels \emph{source} points lying at a minimum average distance from the unlabelled \emph{target} pool. 
%     \item \texttt{UNCERTAINTY} \cite{lewis1995sequential}: labels the most uncertain \emph{source} points, regardless of target data distribution.
%     \item \texttt{KNN-UNCERTAINTY}: labels the most uncertain \emph{source} points lying close to the \emph{target} data, to increase certainty in the neighborhood of target data points.
% \end{enumerate}
% A visualization of the above can be found in \reffig{fig:combined_figures}. 

All strategies build upon reliable distance and uncertainty measures, whose implementation varies based on the type of task, i.e. whether the task is token-level, sequence-level or question answering. A detailed visualization of how these are calculated can be found in \S\ref{app:dist_unc}. Below, we formally describe the three strategies, also detailing the motivation behind our choices. 

\subsection{\texttt{AVERAGE-DIST}} 
\begin{figure*}[htbp]
  \centering
  \begin{subfigure}[b]{0.32\textwidth}
    \includegraphics[width=\textwidth]{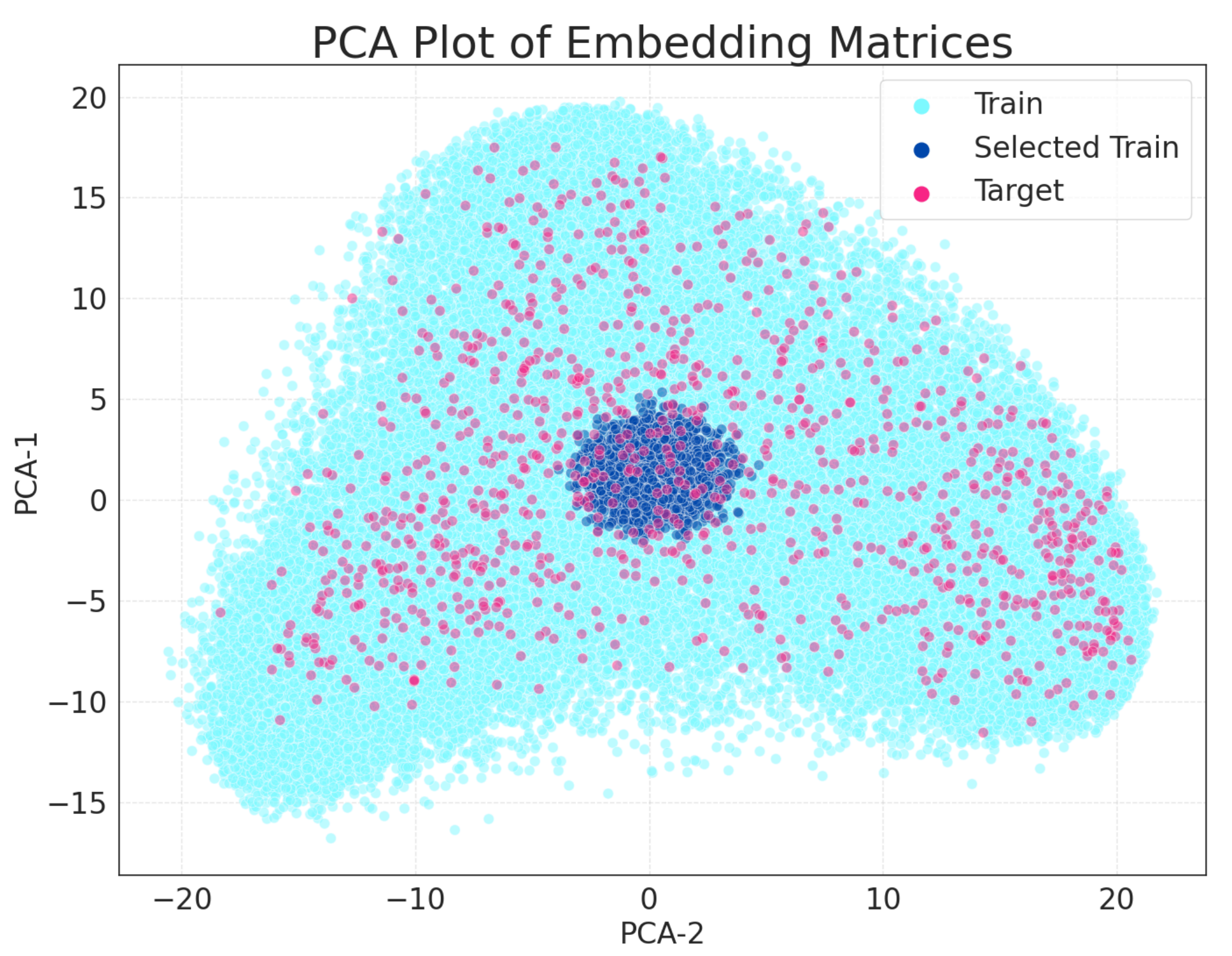}
    \caption{\texttt{AVERAGE-DIST}}
    \label{fig:cls-dist}
  \end{subfigure}
  \hfill
  \begin{subfigure}[b]{0.32\textwidth}
    \includegraphics[width=\textwidth]{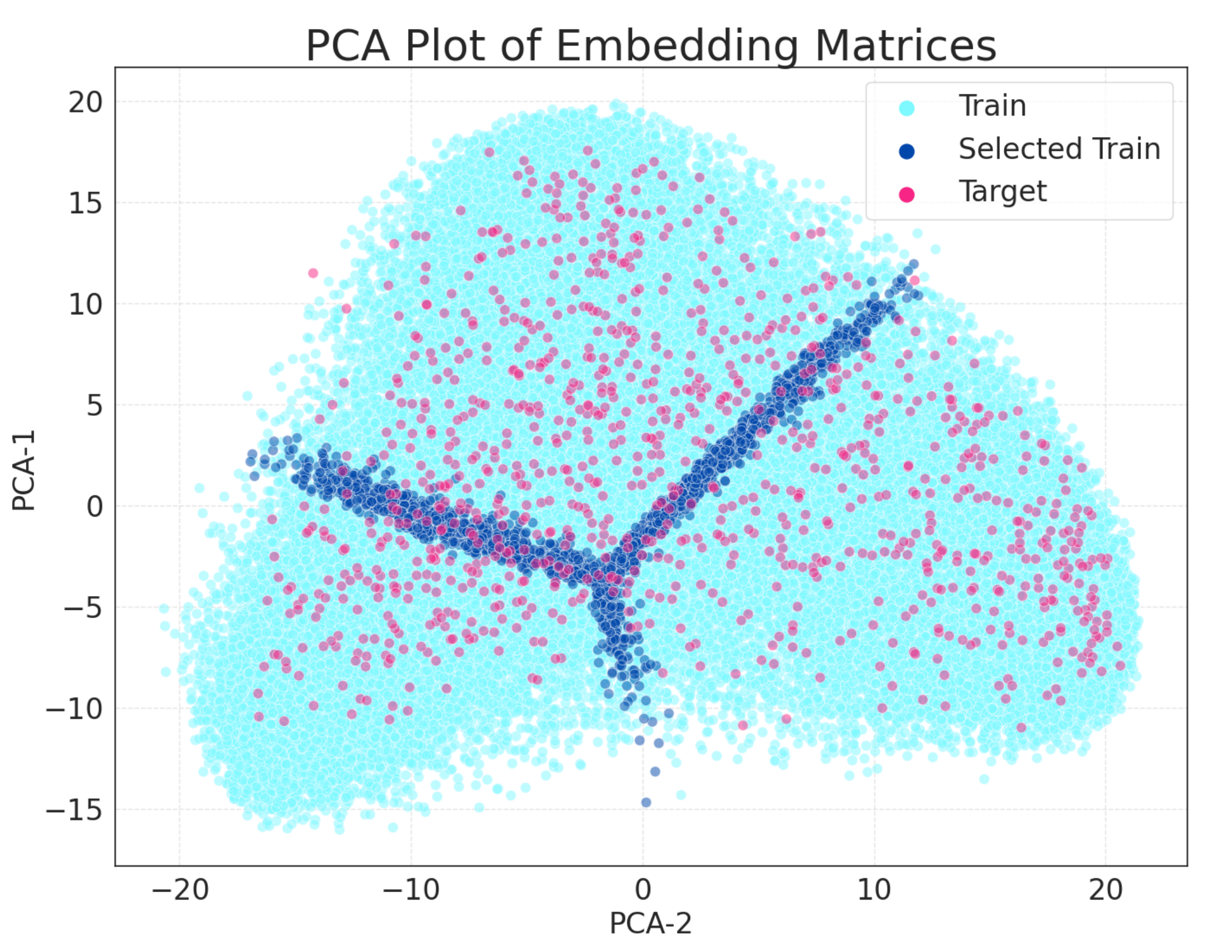}
    \caption{\texttt{UNCERTAINTY}}
    \label{fig:uncertainty}
  \end{subfigure}
  \hfill
  \begin{subfigure}[b]{0.32\textwidth}
    \includegraphics[width=\textwidth]{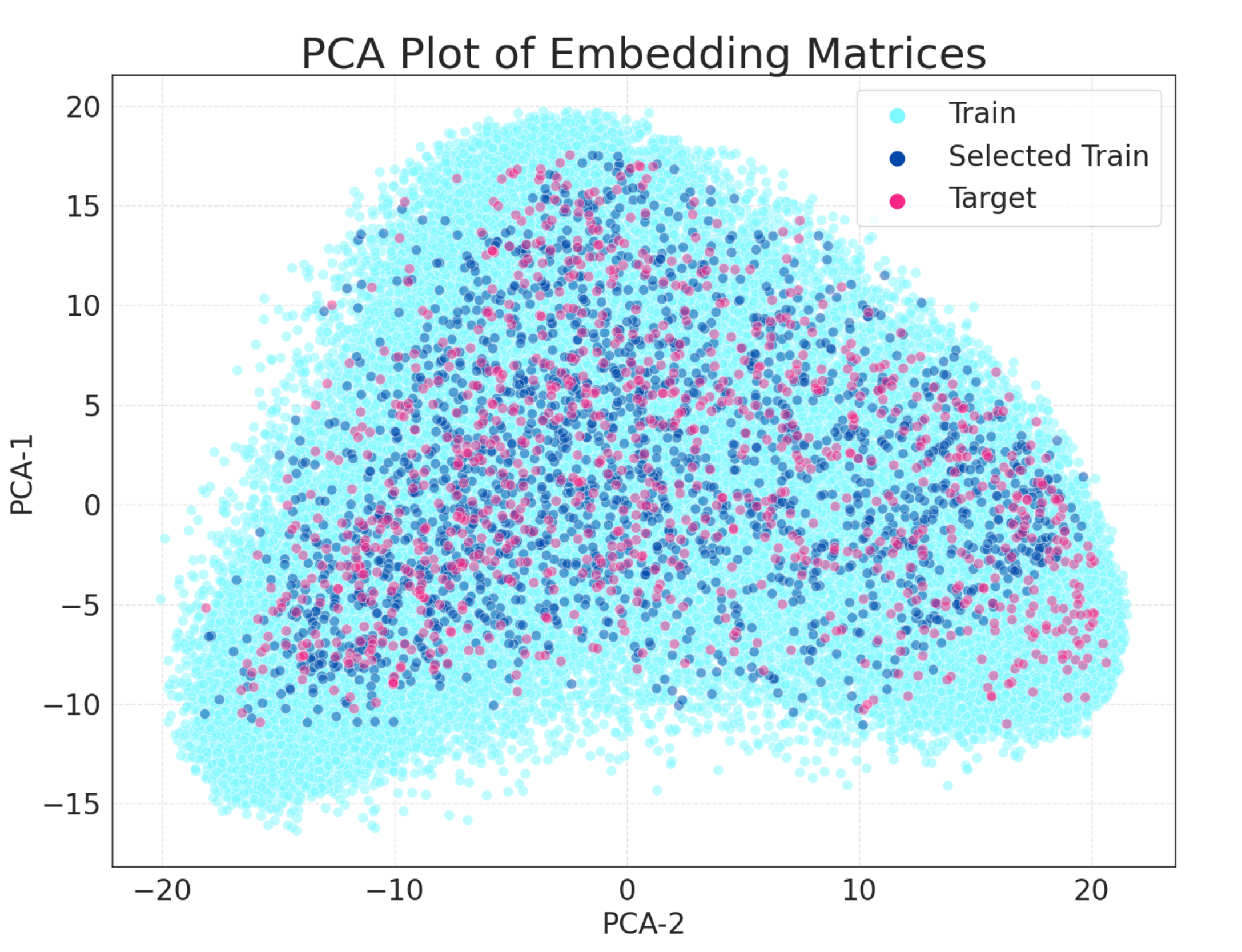}
    \caption{\texttt{KNN-UNCERTAINTY}}
    \label{fig:knn-cls}
  \end{subfigure}
  \caption{\emph{Visualization of datapoints selected using strategies detailed in \refsec{sec:strategies}}, for a three-class sequence classification task (XNLI). \texttt{AVERAGE-DIST} selects points (dark blue) at a minimum average distance from the target (pink); \texttt{UNCERTAINTY} selects most uncertain points lying at the decision boundary of two classes, and \texttt{KNN-UNCERTAINTY} selects uncertain points in the target neighborhood.}
  \label{fig:combined_figures}
\end{figure*}

% \gn{This sub-section doesn't really explain the motivation, maybe add a brief mention?}
\texttt{AVERAGE-DIST} constructs the set $\mathbf{X}^{*}$ such that it minimizes the average distance of points from $\mathcal{X}_{t}$ under an embedding function $f ~:~ \mathcal{X} \rightarrow \mathcal{R}^{d} $ defined by the MultiLM.
This is a representation-based strategy that picks points lying close to the unlabelled target pool \cite{mccallum1998employing,settles2008analysis}. The source points chosen are informative since they are prototypical of the target data in the representation space (Figure \ref{fig:cls-dist}). Especially for low degrees of overlap between source and target data distributions, this criterion can ignore uninformative source points. Formally, 
\begin{gather*}
    \mathbf{X}^{*} = \mathrm{argmin}_{~\mathbf{X} ~\in~ \mathcal{S}^{b}} ~ \sum_{x_{s} \in \mathbf{X}} d_{t}\left(x_{s}\right) \\
    \text{Where}\\
    d_{t}\left(x\right) = \frac{1}{\lvert \mathcal{X}_{t}\rvert} \sum_{x^{j}_{t} \in \mathcal{X}_{t}} \left\| f\left(x\right) -  f\left(x^{j}_{t}\right)\right\|
\end{gather*}

For all task types, we use embeddings of tokens fed into the final classifier, to represent the whole sequence. For NLI and QA, this is the \texttt{[CLS]} token embedding. For token-level tasks, we compute the mean of the initial sub-word token embeddings for each word, as this is the input provided to the classifier to determine the word-level tag.

\subsection{\texttt{UNCERTAINTY}}
\label{sec:uncertainty}
% \gn{this is the first mention of the tasks, and feels a bit sudden. it might be nice to mention somewhere before this what tasks we're handling}\simi{addressed}
Uncertainty sampling \cite{lewis1995sequential} improves annotation efficiency by choosing points that the model would potentially misclassify in the current AL iteration. Uncertainty measures for each task-type can be found below:

\emph{Sequence-Level}: We use margin-sampling \cite{scheffer2001active,schein2007active}, which selects points having the least difference between the model's probabilities for the top-two classes. We compute the output probability distribution for all unlabeled samples in $\mathcal{X}_{s}$ and select samples with the smallest margin. Formally,
\begin{gather*}
    \mathbf{X}^{*} = \mathrm{argmin}_{~\mathbf{X} ~\in~ \mathcal{S}^{b}} ~ \sum_{x_{s} \in \mathbf{X}} P_{\Delta}\left(x_{s}\right) \\
    \text{Where}\\
    P_{\Delta}\left(x\right) = p_{c_1}(x) - p_{c_2}(x)
\end{gather*}

$p_{c_1}(x)$ and $p_{c_2}(x)$ are the predicted probabilities of the top-two classes for an unlabeled sample $x$. 

\vspace{1mm}
% \simi{LD: could you check the first line and rephrase if its tough to understand?}\\
% \ld{I would honestly remove it and put that detail in the appendix}\\
% Here, we experiment with both entropy-based \cite{shen2017deep} and margin-based techniques, but choose the latter because it provides for a higher correlation between the uncertainty of a point and its neighborhood (details in \todo{appendix}). Specifically, f
\emph{Token-level}:  For token-level tasks we first compute the margin (as described above) for each token in the sequence. Then, we assign the minimum margin across all tokens as the sequence margin score and choose  construct $\mathbf{X}^{*}$ with sequences having the least score. Formally,
\begin{gather*}
     \mathbf{X}^{*} = \mathrm{argmin}_{~\mathbf{X} ~\in~ \mathcal{S}^{b}} ~ \sum_{x_{s} \in \mathbf{X}}\texttt{MARGIN-MIN}\left(x_{s}\right) 
    \\
    \text{Where}\\
    \texttt{MARGIN-MIN}\left(x\right) = \min_{i = 1}^{\lvert x \rvert} \left( p^{i}_{c_1}(x) - p^{i}_{c_2}(x) \right)
\end{gather*}

% Above, $p^{i}_{c_1}(x)$ and $p^{i}_{c_2}(x)$ represent the predicted probabilities of the two most probable classes for the $i$th token of unlabeled sample $x$. 

\emph{Question Answering}: The QA task we investigate involves extracting the answer span from a relevant context for a given question. This is achieved by selecting tokens with the highest start and end probabilities as the boundaries, and predicting tokens within this range as the answer. Hence, samples having the lowest start and end probabilities, qualify as most uncertain. Formally, 
\begin{gather*}
     \mathbf{X}^{*} = \mathrm{argmin}_{~\mathbf{X} ~\in~ \mathcal{S}^{b}} ~ \sum_{x_{s} \in \mathbf{X}}\texttt{SUM-PROB}\left(x_{s}\right) 
    \\
    \text{Where}\\
    \texttt{SUM-PROB}\left(x\right) = \max_{i = 1}^{\lvert x \rvert} \log p^{i}_{s}(x) {+} \max_{i = 1}^{\lvert x \rvert} \log p^{i}_{e}(x)
\end{gather*}

Above, $\lvert x \rvert$ denotes the sequence length of the unlabeled sample $x$, and $p^{i}_{s}(x)$ and $p^{i}_{e}(x)$ represent the predicted probabilities for the start  and end index, respectively.

\begin{table*}[ht]
\centering
\small
\begin{tabular}{c | C{2.5cm} | C{3.5cm} | C{5.5cm}}
\toprule
\textbf{Task Type} & \textbf{Task} & \textbf{Dataset} & \textbf{Languages (two-letter ISO code)} \\ \midrule
\multirow{2}{*}{Token-level} & \emph{Part-of-Speech Tagging} (POS) & Universal Dependencies v2.5 \cite{nivre2020universal} & tl, af, ru, nl, it, de, es, bg, pt, fr, te, et, el, fi, hu, mr, kk, hi, tr, eu, id, fa, ur, he, ar, ta, vi, ko, th, zh, yo, ja \\ \cmidrule{2-4}
& \emph{Named Entity Recognition} (NER) & \texttt{WikiAnn} \cite{rahimi2019massively} & nl, pt, bg, it, fr, hu, es, el, vi, fi, et, af, bn, de, tr, tl, hi, ka, sw, ru, mr, ml, jv, fa, eu, ko, ta, ms, he, ur, kk, te, my, ar, id, yo, zh, ja, th \\ \midrule
{\centering Sequence-Level} & \emph{Natural Language Inference} (NLI) & XNLI \cite{conneau2018xnli} & es, bg, de, fr, el, vi, ru, zh, tr, th, ar, hi, ur, sw \\ \midrule
\multicolumn{2}{c|}{Question Answering (QA)} & TyDiQA \cite{clark2020tydi} & id, fi, te, ar, ru, sw, bn, ko \\ 
\bottomrule
\end{tabular}
\caption{\emph{Tasks and Datastes}: \method{} is applied across tasks of varying complexity, as elucidated in \textbf{Q1}: \S\ref{sec:experiments}.}
\label{tab:dataset_stats}
\end{table*}

\subsection{\texttt{KNN-UNCERTAINTY}}
As standalone measures, both distance and uncertainty based criteria have shortcomings. When there is little overlap between source and target, choosing source points based on \texttt{UNCERTAINTY} alone leads to selecting data that are uninformative to the target. When there is high degrees of overlap between source and target, the \texttt{AVERAGE-DIST} metric tends to produce a highly concentrated set of points (Figure \ref{fig:cls-dist}) -- even if the model is accurate in that region of representation space -- resulting in minimal coverage on the target set.

To design a strategy that combines the strengths of both distance and uncertainty, we first measure how well a target point's uncertainty correlates with its neighborhood. We calculate the Pearson's correlation coefficient ($\rho$) \cite{pearson1903mathematical} between the uncertainty of a target point in $\mathcal{X}_{t}$ and the average uncertainty of its top-$k$ neighbors in $\mathcal{X}_{s}$. We observe a statistically significant $\rho$ value > 0.7, for all tasks. A natural conclusion drawn from this is that decreasing the uncertainty of a target point's neighborhood would decrease the uncertainty of the target point itself. Hence, we first select the top-$k$ neighbors for each $x_t \in \mathcal{X}_{t}$. Next, we choose the most uncertain points from these neighbors until we reach $b$ data points. Formally, until $\lvert \mathbf{X}^{*} \rvert =b$ :
% \vspace{-2mm}
\begin{gather*}
     \mathbf{X}^{*} = \mathrm{argmax}_{\{\mathbf{X} ~\subset~ \mathcal{N}_{t}^{k} ~\big|~ \lvert \mathbf{X} \rvert ~=~ b \}} \sum_{x_s ~\in~ \mathbf{X}} U(x_s)
    \\
    \text{Where}\\
    \mathcal{N}_{t}^{k} = \bigcup_{j=1}^{|\mathcal{X}_t|} ~\texttt{k-NEARESTNEIGHBORS}(x^j_t, \mathcal{X}_{s})
\end{gather*}

% \begin{equation}
% D_{src}^{l} = \arg\max_{d_{src}^{ul}} U(d_{src}^{ul}) \ \forall \ d_{src}^{ul} \in \bigcup_{j=1}^{|D_{tgt}^{ul}|} N(d_{tgt_j}^{ul}, k)
% \end{equation}
Above, $U(x_s)$ represents the uncertainty of the source point as calculated in \S\ref{sec:uncertainty}.

\section{Experimental Setup}
\label{sec:experiments}

\begin{table*}[ht]
\centering
\small
\begin{tabular}{c | c | c | c | C{3.4cm} | C{5.4cm}}
\toprule
\multirow{2}{*}{\textbf{Dataset}} & \multicolumn{3}{c}{\textbf{Single Target}} & \multicolumn{2}{c}{\textbf{Multi-Target}} \\
\cmidrule(lr){2-4} \cmidrule(lr){5-6}
& \textbf{HP} & \textbf{MP} & \textbf{LP} & \textbf{Geo} & \textbf{LPP} \\ \midrule
UDPOS & French & Turkish & Urdu & Telugu, Marathi, Urdu & Arabic, Hebrew, Japanese, Korean, Chinese, Persian, Tamil, Vietnamese, Urdu \\ \midrule
NER & French & Turkish & Urdu & Indonesian, Malay, Vietnamese & Arabic, Indonesian, Malay, Hebrew, Japanese, Kazakh, Malay, Tamil, Telugu, Thai, Yoruba, Chinese, Urdu \\ \midrule
XNLI & French & Turkish & Urdu & Bulgarian, Greek, Turkish & Arabic, Thai, Swahili, Urdu, Hindi \\ \midrule
TyDiQA & Finnish & Arabic & Bengali & Bengali, Telugu & Swahili, Bengali, Korean \\
\bottomrule
\end{tabular}
\caption{\emph{Target language configurations}. We run five experiments for each model and task, with the language sets above as targets (details in \S\ref{sec:lang_sel}). All languages mentioned in Table \ref{tab:dataset_stats} make up the source set, \textbf{except} the chosen target languages for a particular configuration.}
\label{tab:langs}
\end{table*}

Our setup design aims to address the following: 

\vspace{1mm}
\noindent \textbf{Q1)} Does \method{} benefit tasks with varying complexity? Which strategies work well across different task types? (\S\ref{sec:tasks}) \\
\noindent \textbf{Q2)} How well does \method{} perform across a varied set of target languages? Can it benefit multilingual target pools as well? (\S\ref{sec:lang_sel}) \\
\noindent \textbf{Q3)} How do the benefits of \method{} vary across different MultiLMs? (\S\ref{sec:models})

\subsection{Task and Dataset Selection}
\label{sec:tasks}
We have three distinct task types, based on the label format. We remove duplicates from each dataset to prevent selecting multiple copies of the same instance. Dataset details can be found in Table \ref{tab:dataset_stats}.

\subsection{Source and Target Language Selection}
\label{sec:lang_sel}
We experiment with the zero-shot case of disjoint source and target languages, i.e., the unlabelled source pool contains no data from target languages. The train and validation splits constitute the unlabelled source or target data, respectively. Evaluation is done on the test split for each target language. With \textbf{Q2)} in mind, we experiment with five target settings (Table \ref{tab:langs}): 

\vspace{1mm}
\textbf{Single-target}: We partition languages into three equal tiers based on zero-shot performance post fine-tuning on English: high-performing (\emph{HP}), mid-performing (\emph{MP}) and low-performing (\emph{LP}), and choose one language from each, guided by two factors. First, we select languages that are common across multiple datasets, to study how data selection for the same language varies across tasks. From these, we choose languages that have similarities with the source set across different linguistic dimensions (obtained using \texttt{lang2vec} \cite{littell2017uriel}), to study the role of typological similarity for different tasks.

\vspace{1mm}
\textbf{Multi-target}: Here, we envision two scenarios: 

\emph{Geo}: Mid-to-low performing languages in geographical proximity are chosen. From an application perspective, this would allow one to improve a MultiLM for an entire geographical area.

\emph{LPP}: All low-performing languages are pooled, to test whether we can collectively enhance the MultiLM's performance across all of them.

% \vspace{1mm}
% \textbf{Token-Level}: Each word in the sequence is associated with a single label. Tasks here include: 

% \emph{Named Entity Recognition} (NER): For NER, we use the \texttt{WikiAnn} dataset \cite{rahimi2019massively} which consists of named entities in Wikipedia, automatically annotated with \texttt{LOC}, \texttt{PER} and \texttt{ORG} tags in IOB2 format.

% \emph{Part-of-Speech Tagging} (UDPOS): We use the Universal Dependencies v2.5 \cite{nivre2020universal} treebanks for POS tagging. Each word is assigned one of 17 universal POS tags.  

% \vspace{1mm}
% \textbf{Sequence-Level}: Each sequence is associated with a single label. Tasks here include:

% \emph{Natural Language Inference} (NLI): The task is to determine whether a premise entails, contradicts, or is neutral toward a hypothesis. We use the XNLI \cite{conneau2018xnli} dataset, featuring manually translated validation and test sets, and machine-translated training sets, using a subset of MultiNLI \cite{williams2017broad}.

% \vspace{1mm}
% \textbf{Question Answering} (QA): We use the gold passage version of the TyDiQA dataset \cite{clark2020tydi}, that covers 11 typologically diverse languages, and is curated from Wikipedia.

\subsection{Model Selection}
\label{sec:models}
We test \method{} across multiple MultiLMs: \emph{XLM-R} \cite{conneau2019unsupervised}, \emph{InfoXLM} \cite{chi2020infoxlm}, and \emph{RemBERT} \cite{chung2020rethinking}. All models have a similar number of parameters ($\sim$550M-600M), and support 100+ languages. \emph{XLM-R} is trained on monolingual corpora from CC-100 \cite{conneau2019unsupervised}, \emph{InfoXLM} is trained to maximize mutual information between multilingual texts, and \emph{RemBERT} is a deeper model, that reallocates input embedding parameters to the Transformer layers. 
% \emph{InfoXLM} and \emph{RemBERT} report improved performance over \emph{XLM-R}. 

\subsection{Baselines} 
We include a number of baselines to compare our strategies against: 

\vspace{1mm}
\noindent \textbf{1)} \texttt{RANDOM}: In each round, a random subset of $b$ data points from $\mathcal{X}_s$ is selected.

\noindent \textbf{2)} \texttt{EGALITARIAN}: An equal number of randomly selected data points from the unlabeled pool for each language, i.e. $|x_{s}|=b/|L_{s}|$; $\forall x_{s} \in \mathcal{X}_{s}$ is chosen. \citet{debnath2021towards} demonstrate that this outperforms a diverse set of alternatives.

\noindent \textbf{3)} \texttt{LITMUS}: \texttt{LITMUS} \cite{srinivasan2022litmus} is a tool to make performance projections for a fine-tuned model, but can also be used to generate data labeling plans, based on the predictor's projections. We only run this for XLM-R since the tool requires past fine-tuning performance profiles, and XLM-R is supported by default.

\noindent \textbf{4)} \texttt{GOLD}: This involves training on data from the target languages itself. Given all other strategies are zero-shot, we expect \texttt{GOLD} to out-perform them and help determine an upper bound on performance.

\subsection{Fine-tuning Details} 
We first fine-tune all MultiLMs on English (\texttt {EN-FT}) and continue fine-tuning on data selected using \method, similar to \citet{lauscher2020zero,kumar2022diversity}. We experiment with a budget of 10,000 examples acquired in five AL rounds, except for TyDiQA, where our budget is 5,000 examples (TyDiQA is of the order of 35-40k samples overall across ten languages, and this is to ensure fair comparison with our gold strategy). For each model, we first obtain \texttt {EN-FT} and continually fine-tune using \method{}. Hyperparameter details are given in \S\ref{app:ft}, and all results are averaged across three seeds: 2, 22, 42. We fine-tune using a fixed number of epochs without early stopping, given the lack of a validation set in our setup (we assume no labelled target data).

\section{Results}
\label{sec:results}

\textbf{How  does \method{} perform overall?} We present results for NER, POS, NLI and QA in Tables \ref{tab:panx}, \ref{tab:udpos}, \ref{tab:xnli} and \ref{tab:tydiqa}, respectively. In summary, the best-performing strategies outperform the best performing baselines in 84\% of the cases, with variable gains dependant on the task, model and target languages. In the remaining cases, the drop is within 1\% absolute delta from the best-performing baseline. 

\vspace{1mm}
\noindent \textbf{How does \method{} fare on multilingual target pools?} We observe consistent gains given multilingual target pools as well (\emph{Geo} and \emph{LPP}). We believe this is enabled by the language-independent design of our strategies, which makes annotation decisions at a per-instance level. This has important consequences, since this would enable researchers, like those at Company \textbf{Y}, to better models for all the languages that they care about. 

\vspace{1mm}
\noindent \textbf{Does the model select data from the same languages across tasks?} No! We find that selected data distributions vary across tasks for the same target languages. For example, when the target language is Urdu, \method{} chooses 70-80\% of samples from Hindi for NLI and POS, but prioritizes Farsi and Arabic (35-45\%) for NER. Despite Hindi and Urdu's syntactic, genetic, and phonological similarities as per \texttt{lang2vec}, their differing scripts underscore the significance of script similarity in NER transfer. This also proves that analysing data selected by \method{} can offer linguistic insights into the learned task-specific representations.

\begin{table}[ht]
\centering
\small
\begin{tabular}{@{}c@{ }|c|c|c|c|c|c}
\toprule
\multicolumn{1}{c}{} & \textbf{Method} & \textbf{HP} & \textbf{MP} & \textbf{LP} & \textbf{Geo} & \textbf{LPP} \\ 
\midrule
\multirow{6}{*}{\rotatebox[origin=c]{90}{XLM-R}} 
& \texttt{EN-FT} & 80.0 & 79.5 & 65.6 & 61.0 & 45.8 \\
& \texttt{GOLD} & 90.1 & 92.8 & 94.5& 81.2& 73.7\\
& \texttt{BASE}$_{egal}$ & 85.4 & 87.6& 84.0& 80.6& 62.8\\
& $\method_{knn}$ & 87.8 & 89.2& 85.8& 82.4& 62.3\\
\cmidrule{2-7}
& $\Delta_{base}$ & \textbf{2.4} & \textbf{1.6}& \textbf{1.8}& \textbf{1.8}& -0.5\\
\midrule
\multirow{6}{*}{\rotatebox[origin=c]{90}{InfoXLM}} 
& \texttt{EN-FT} & 80.5 & 82.8 & 65.4 & 64.2 & 44.8 \\
& \texttt{GOLD} & 90.0& 92.8& 94.6& 83.5& 74.9\\
& \texttt{BASE}$_{egal}$ & 84.0& 87.6& 83.2& 80.9& 63.4\\
& $\method_{knn}$ & 87.4& 89.2& 85.5& 82.2& 64.2\\
\cmidrule{2-7}
& $\Delta_{base}$ & \textbf{3.4}& \textbf{1.6}& \textbf{2.2} & \textbf{1.3}& \textbf{0.8}\\
\midrule
\multirow{6}{*}{\rotatebox[origin=c]{90}{RemBERT}} 
& \texttt{EN-FT} & 78.8 & 80.2 & 55.7 & 61.1 & 48.4 \\
& \texttt{GOLD} & 89.4 & 92.1& 93.5 & 79.8& 70.1 \\
& \texttt{BASE}$_{egal}$ & 84.6 & 86.8& 82.3& 79.2& 59.8 \\
& $\method_{knn}$ & 87.1 & 89.0& 85.7& 79.8 & 62.1 \\
\cmidrule{2-7}
& $\Delta_{base}$ & \textbf{2.5} & \textbf{2.2}& \textbf{3.4}& \textbf{0.6}& \textbf{2.3} \\
\bottomrule
\end{tabular}
\caption{ \emph{PAN-X Results (F1)}: We observe gains across all models and \texttt{KNN-UNCERTAINTY} performs best. $\Delta_{base}$ represents the delta from baseline.}
\label{tab:panx}
\end{table}

\begin{table}[ht]
\centering
\small
\begin{tabular}{@{}c@{ }|c|c|c|c|c|c}
\toprule
\multicolumn{1}{c}{} & \textbf{Method} & \textbf{HP} & \textbf{MP} & \textbf{LP} & \textbf{Geo} & \textbf{LPP} \\ 
\midrule
\multirow{5}{*}{\rotatebox[origin=c]{90}{XLM-R}} 
& \texttt{EN-FT} & 81.7 & 75.5 & 71.5 & 80.2 & 62.2 \\
& \texttt{GOLD} & 95.6 & 81.2 & 93.2 & 91.8 & 88.2 \\
& \texttt{BASE}$_{egal}$ & 87.1 & 79.6 & 88.4 & 85.7 & 68.9 \\
& $\method_{knn}$ & 87.5 & 80.1 & 90.1 & 86.1 & 70.9 \\
\cmidrule{2-7}
& $\Delta_{base}$ & \textbf{0.4} & \textbf{0.5} & \textbf{1.7} & \textbf{0.4} & \textbf{2.0} \\
\midrule
\multirow{5}{*}{\rotatebox[origin=c]{90}{InfoXLM}} & \texttt{EN-FT} & 79.6 & 74.0 & 59.0 & 73.6 & 58.2 \\
& \texttt{GOLD} & 95.7 & 81.4 & 93.3 & 92.0 & 88.7 \\
& \texttt{BASE}$_{egal}$ & 88.0 & 79.4 & 88.8 & 86.3 & 67.8 \\
& $\method_{knn}$ & 87.8 & 79.5 & 90.4 & 86.0 & 66.8 \\
\cmidrule{2-7}
& $\Delta_{base}$ & -0.3 & \textbf{0.1} & \textbf{1.6} & -0.3 & -1.0 \\
\midrule
\multirow{5}{*}{\rotatebox[origin=c]{90}{RemBERT}} 
& \texttt{EN-FT} & 72.9 & 71.1 & 50.6 & 66.1 & 55.7 \\
& \texttt{GOLD} & 95.1 & 80.8 & 92.3 & 91.2 & 86.8 \\
& \texttt{BASE}$_{egal}$ & 86.9 & 78.1 & 85.8 & 83.8 & 67.8 \\
& $\method_{knn}$ & 87.4 & 77.7 & 88.2 & 84.2 & 68.0 \\
\cmidrule{2-7}
& $\Delta_{base}$ & \textbf{0.5} & -0.3 & \textbf{2.4} & \textbf{0.4} & \textbf{0.2} \\
\bottomrule
\end{tabular}
\caption{\emph{UDPOS Results (F1)}: We observe modest gains for a 10k budget, but higher gains for lower budgets (\S\ref{sec:analysis})}.
\label{tab:udpos}
\end{table}

\begin{table}[ht]
\centering
\small
\begin{tabular}{@{}c@{ }|c|c|c|c|c|c}
\toprule
\multicolumn{1}{c}{} & \textbf{Method} & \textbf{HP} & \textbf{MP} & \textbf{LP} & \textbf{Geo} & \textbf{LPP} \\ 
\midrule
\multirow{6}{*}{\rotatebox[origin=c]{90}{XLM-R}} & \texttt{EN-FT} & 81.8 & 77.3 & 69.9 & 80.1 & 73.4 \\
& \texttt{GOLD} & 81.6 & 79.5 & 70.3 & 81.6 & 76.0 \\
& \texttt{BASE}$_{egal}$ & 81.6 & 78.8 & 73.0 & 80.9 & 75.6 \\
& $\method_{avg}$ & 83.7 & 79.9 & 75.3 & 82.2 & 77.1 \\
\cmidrule{2-7}
& $\Delta_{base}$ & \textbf{2.1} & \textbf{1.1} & \textbf{2.3} & \textbf{1.3} & \textbf{1.5} \\
& $\Delta_{gold}$ & \textbf{2.1} & \textbf{0.4} & \textbf{5.0} & \textbf{0.6} & \textbf{1.1} \\
\midrule
\multirow{6}{*}{\rotatebox[origin=c]{90}{InfoXLM}} & \texttt{EN-FT} & 81.9 & 77.3 & 68.8 & 79.8 & 71.5 \\
& \texttt{GOLD} & 83.6 & 80.6 & 73.7 & 82.4 & 77.7 \\
& \texttt{BASE}$_{egal}$ & 83.7 & 79.8 & 74.6 & 81.5 & 77.3 \\
& $\method_{avg}$ & 84.8 & 80.8 & 75.9 & 83.1 & 77.8 \\
\cmidrule{2-7}
& $\Delta_{base}$ & \textbf{1.1} & \textbf{1.0} & \textbf{1.3} & \textbf{1.6} & \textbf{0.5} \\
& $\Delta_{gold}$ & \textbf{1.2} & \textbf{0.2} & \textbf{2.2} & \textbf{0.7} & \textbf{0.1} \\
\midrule
\multirow{6}{*}{\rotatebox[origin=c]{90}{RemBERT}} & \texttt{EN-FT} & 83.1 & 73.9 & 52.5 & 77.8 & 63.0 \\
& \texttt{GOLD} & 81.1 & 73.3 & 63.1 & 76.0 & 67.5 \\
& \texttt{BASE}$_{egal}$ & 80.0 & 75.3 & 63.9 & 76.4 & 67.9\\
& $\method_{avg}$ & 81.7 & 76.1 & 67.6 & 78.6 & 70.9 \\
\cmidrule{2-7}
& $\Delta_{base}$ & \textbf{1.7} & \textbf{0.8} & \textbf{3.7} & \textbf{2.2} & \textbf{3.0} \\
& $\Delta_{gold}$ & \textbf{0.6} & \textbf{2.8} & \textbf{4.5} & \textbf{2.6} & \textbf{3.4} \\
\bottomrule
\end{tabular}
\caption{\emph{XNLI Results (F1)}: Here we even surpass the gold standard of in-language finetuning. Details in \S\ref{sec:results}.} 
\label{tab:xnli}
\end{table}

\begin{table}[ht]
\centering
\small
\begin{tabular}{@{}c@{ }|c|c|c|c|c|c}
\toprule
\multicolumn{1}{c}{} & \textbf{Method} & \textbf{HP} & \textbf{MP} & \textbf{LP} & \textbf{Geo} & \textbf{LPP} \\ 
\midrule
\multirow{5}{*}{\rotatebox[origin=c]{90}{XLM-R}} & \texttt{EN-FT} & 78.9 & 73.2 & 79.9 & 80.7 & 78.5 \\
& \texttt{GOLD} & 81.2& 83.8& 83.7& 84.7& 81.0\\
& \texttt{BASE}$_{egal}$ & 79.9& 81.7& 79.6& 81.1& 78.7 \\
& $\method_{unc}$ & 80.8& 82.9& 80.3& 81.0& 77.8\\
\cmidrule{2-7}
& $\Delta_{base}$ & \textbf{0.9}& \textbf{1.2}& \textbf{0.7}& -0.1& -0.9\\
\midrule
\multirow{5}{*}{\rotatebox[origin=c]{90}{InfoXLM}} 
& \texttt{EN-FT} & 77.6 & 75.4 & 82.2 & 81.9 & 78.8 \\
& \texttt{GOLD} & 80.7& 85.0& 86.3& 87.1& 81.1\\
& \texttt{BASE}$_{egal}$ & 80.5& 82.3& 82.2& 82.3& 78.1\\
& $\method_{unc}$ & 81.8& 84.1& 80.8& 82.6& 77.8\\
\cmidrule{2-7}
& $\Delta_{base}$ & \textbf{1.3}& \textbf{1.8}& -1.4& \textbf{0.3}& -0.3\\
\midrule
\multirow{5}{*}{\rotatebox[origin=c]{90}{RemBERT}} & \texttt{EN-FT} & 79.7 & 73.0 & 82.9 & 78.0 & 78.2 \\
& \texttt{GOLD} & 78.4& 80.1& 86.7& 84.4& 80.5\\
& \texttt{BASE}$_{egal}$ & 81.3& 78.9& 82.8& 76.5& 75.3\\
& $\method_{unc}$ & 82.7& 80.2& 80.6& 78.0& 76.1\\
\cmidrule{2-7}
& $\Delta_{base}$ & \textbf{1.4}& \textbf{1.3}& -2.2& \textbf{1.5}& \textbf{0.8}\\
\bottomrule
\end{tabular}
\caption{\emph{TyDiQA Results (F1)}: \texttt{UNCERTAINTY} works best here. Despite TyDiQA being composed of typologically diverse languages and being extremely small (35-40k samples), we observe modest gains across multiple configs.}
\label{tab:tydiqa}
\end{table}

\begin{figure}[!htbp]
    \includegraphics[width=0.47\textwidth]{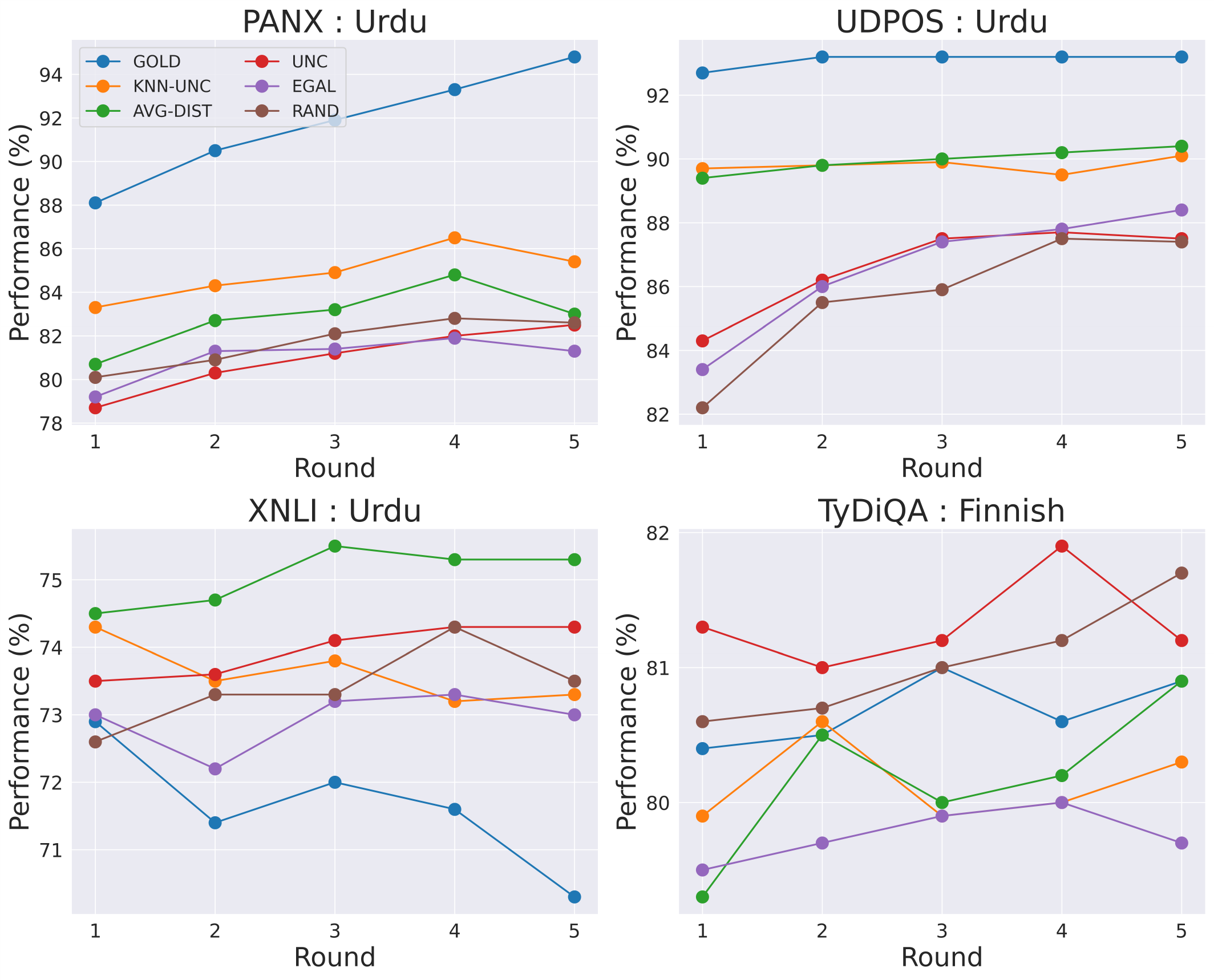}
    \caption{\emph{Performance across different AL rounds for XLM-R}: Semantic similarity with target matters most for token-level tasks, but uncertainty gains importance for NLI and QA. (\texttt{AVG-DIST} picks maximally uncertain points for NLI due to the nature of the dataset (\S\ref{sec:results}).}
    \label{fig:rounds}
\end{figure}

\begin{figure*}[!htbp]
    \includegraphics[width=\textwidth]{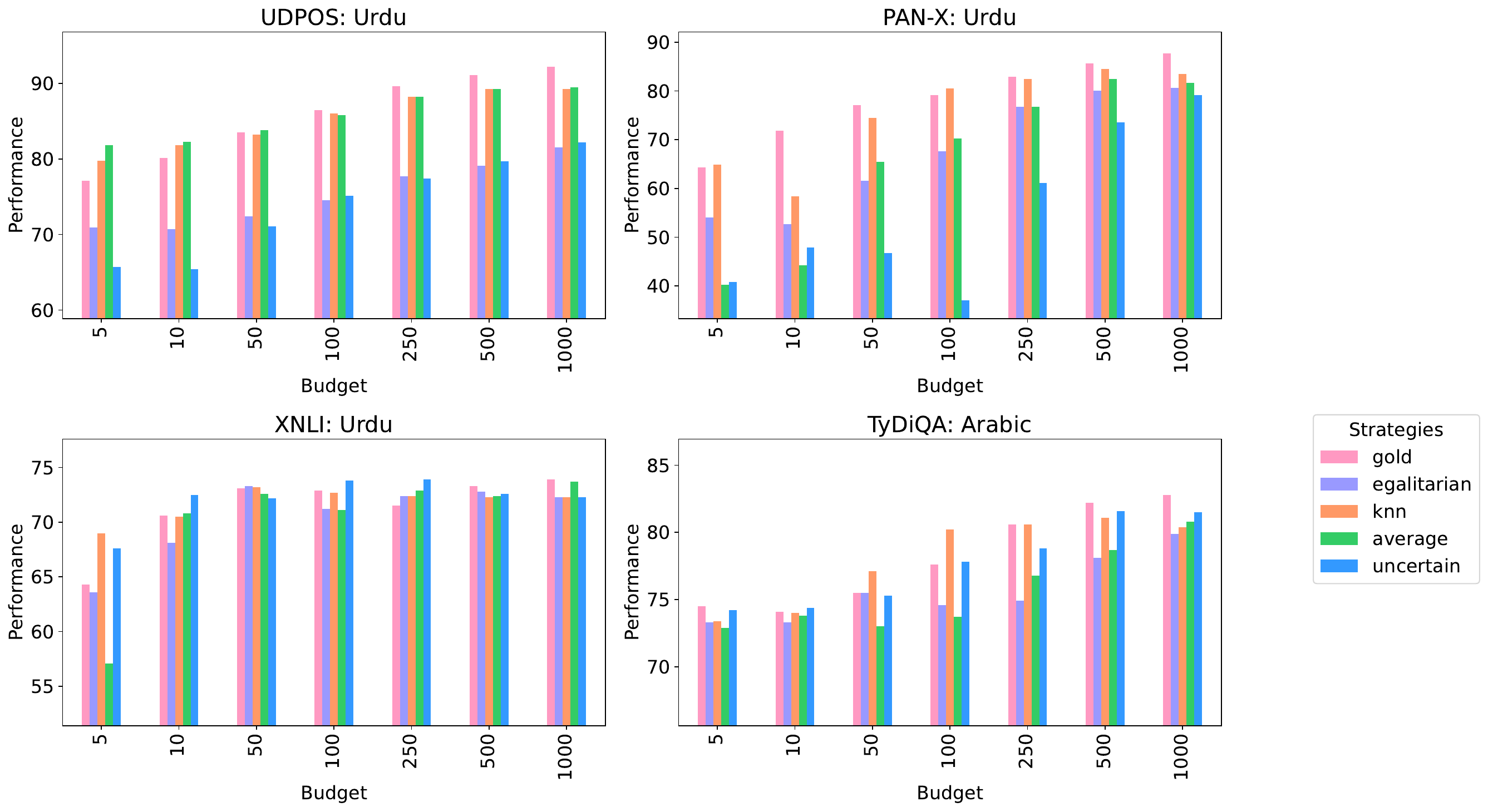}
    % \caption{\emph{Ablation result}: - \gn{Can we make a line chart instead of bar chart? Also, it seems that 5, 50, 100, 250 are missing?}\simi{made a line chart for the setting where we have multiple rounds of fine-tuning. Since this is a single round of fine-tuning and multiple budgets I kept the bar chart, it was also helping visualize the changes better since the randge becomes too large in the line charts}}
    \caption{\emph{Multiple budgets, one AL round}: We experiment with low-budgets acquired using the \texttt{\small EN-FT} model. We observe gains of up to 8-11 F1 over baselines for 5-100 examples, with a trend of diminishing gains given larger budgets. All runs averaged across three seeds (2, 22, 42).}
    \label{fig:ablation}
\end{figure*}

\vspace{1mm}
\noindent \textbf{Which strategies work well across different task types?}
Our hybrid strategy, which picks uncertain points in the local neighborhood of target points, performs best for token-level tasks, whereas globally uncertain points maximize performance for NLI and QA. For NLI, both \texttt{AVERAGE-DIST} and \texttt{UNCERTAINTY} outperform baselines, the former proving more effective. On further analysis, we find that this is an artifact of the the nature of the dataset which is balanced across three labels, and is strictly parallel. This makes \texttt{AVERAGE-DIST} select high-uncertainty points at decision boundaries' centroid, as visualized in Figure \ref{fig:cls-dist}. Our finding of different strategies working well for different tasks is consistent with past works. For example, \citet{settles2008analysis} find information density (identifying semantically similar examples), to work best for sequence-labeling tasks; \citet{marti2022active} discuss how uncertainty-based methods perform best for QA; and \citet{kumar2022diversity} also recommend task-dependant labeling strategies. By including multiple tasks and testing multiple strategies for each, we provide guidance to everyday practitioners on the best strategy, given a task.

\vspace{1mm}

\noindent \textbf{How does performance vary across models?} For all models, the \texttt{EN-FT} performance is similar, except for UDPOS, where XLM-R is consistently better than InfoXLM and RemBERT. We see this play a role in the consistent gains provided by \method across languages for XLM-R, as opposed to RemBERT and InfoXLM. We hypothesize that the better a model's representations, the better our measures of distance and uncertainty, which inherently improves data selection.

\section{Further Analysis}
\label{sec:analysis}
\noindent \textbf{What is the minimum budget for which we can observe gains in one AL round?} To deploy \method{} in resource-constrained settings, we test its applicability in low-budget settings, ranging from 5,10,50,100,250,500,1000, acquired using the \textsc{\small EN-FT} model only. As shown in Figure \ref{fig:ablation}, we observe gains across all budget levels. Notably, we observe gains of up to 8-11 F1 points for token-level tasks, and 2-5 F1 points for NLI and QA, for most lower budgets (5-100 examples). These gains diminish as the budget increases. For complex tasks like NLI and QA, semantic similarity with the target holds importance when the budgets is below 500 examples, but picking globally uncertain points gains precedence for larger budgets. 

\vspace{1mm}
\noindent \textbf{Do the selected datapoints matter or does following the language distribution suffice?} \method{} not only identifies transfer languages but also selects specific data for labeling. To evaluate its importance, we establish the language distribution of data selected using \method{} and randomly select datapoints following this distribution. Despite maintaining this distribution, performance still declines (\S\ref{app:detailed}), indicating that precise datapoint selection in identified transfer languages is vital.

\section{Related Work}
\noindent \textbf{Multilingual Fine-tuning}: Traditionally models were fine-tuned on English, given the availability of labeled data across all tasks. However, significant transfer gaps were observed across languages \cite{hu2020xtreme} leading to the emergence of two research directions. The first emphasizes the significance of using few-shot target language data \cite{lauscher2020zero} and the development of strategies for optimal few-shot selection \cite{kumar2022diversity,moniz2022efficiently}. The second focuses on choosing the best source languages for a target, based on linguistic features \cite{lin2019choosing} or past model performances \cite{srinivasan2022litmus}. Discerning a globally optimal transfer language however, has been largely ambiguous \cite{pelloni2022subword} and the language providing for highest empirical transfer is at times inexplicable by known linguistic relatedness criteria \cite{pelloni2022subword, turc2021revisiting}. By making decisions at a data-instance level rather than a language level, \method \space removes the reliance on linguistic features and sidesteps ambiguous consensus on how MultiLMs learn cross-lingual relations, while prescribing domain-relevant instances to label. \\

\noindent \textbf{Active learning for NLP}: AL has seen wide adoption in NLP, being applied to tasks like text classification \cite{karlos2012empirical,li2013active}, named entity recognition \cite{shen2017deep,wei2019cost,erdmann2019practical}, and machine translation \cite{miura2016selecting,zhao2020active}, among others. In the multilingual context, past works \cite{moniz2022efficiently,kumar2022diversity,chaudhary2019little} have applied AL to selectively label data in target languages. However, they do not consider cases with unknown overlap between source and target languages. This situation, similar to a multi-domain AL setting, is challenging as data selection from the source languages may not prove beneficial for the target \cite{longpre2022active}. 

\section{Conclusion}
\label{sec:conclusion}
In this work, we introduce \method{}, an end-to-end framework that selects data to label from vast pools of unlabelled multilingual data, under an annotation budget. \method{}'s design is language-agnostic, making it viable for cases where source and target data do not overlap. We design three strategies drawing from AL principles that encompass semantic similarity with the target, uncertainty, and a hybrid combination of the two. Our strategies outperform strong baselines for 84\% of target language configurations (including multilingual target sets) in the zero-shot case of disjoint source and target languages, across three models and four tasks: NER, UDPOS, NLI and QA. We find that semantic similarity with the target mostly benefits token-level tasks, while picking uncertain points gains precedence for complex tasks like NLI and QA. We further analyse \method{}'s applicability in low-budget settings and observe gains of up to 8-11 F1 points for some tasks, with a trend of diminishing gains for larger budgets. We hope that our work helps improve the capabilities of MultiLMs for desired languages, in a cost-efficient way.
\section{Limitations}
\label{sec:limitations}
With \method{}'s wider applicability across languages come a few limitations as we detail below: \\

\noindent \textbf{Inference on source data}: \method{} relies on model representations and its output ditribution for each example. This requires us to run inference on all of the source data; which can be time-consuming. However, one can run parallel CPU-inference which greatly reduces latency. \\

\noindent \textbf{Aprior Model Selection}: We require knowing the model apriori which might mean a different labeling scheme for different models.  This a trade-off we choose in pursuit of better performance for the chosen model, but it may not be the most feasible solution for all users. \\

\noindent \textbf{Refinement to the hybrid approach}: Our hybrid strategy picks the most uncertain points in the neighborhood of the target. However, its current design prioritizes semantic similarity with the target over global uncertainty, since we first pick top-$k$ neighbors and prune this set based on uncertainty. However, it will be interesting to experiment with choosing globally uncertain points first and then pruning the set based on target similarity. For NLI and QA, we observe that globally uncertain points help for higher budgets but choosing nearest neighbors helps most for lower budgets. Therefore, this alternative may work better for these tasks, and is something we look to explore in future work.

\section{Acknowledgements}
\label{sec:ack}
We thanks members of the Neulab and COMEDY, for their invaluable feedback on a draft of this paper. This work was supported in part by grants from the National Science Foundation (No.~2040926), Google, Two Sigma, Defence Science and Technology Agency (DSTA) Singapore, and DSO National Laboratories Singapore.

\bibliography{custom}
\bibliographystyle{acl_natbib}

\appendix

\section{Appendix}
\begin{figure*}[!htbp]
    \includegraphics[width=\textwidth]{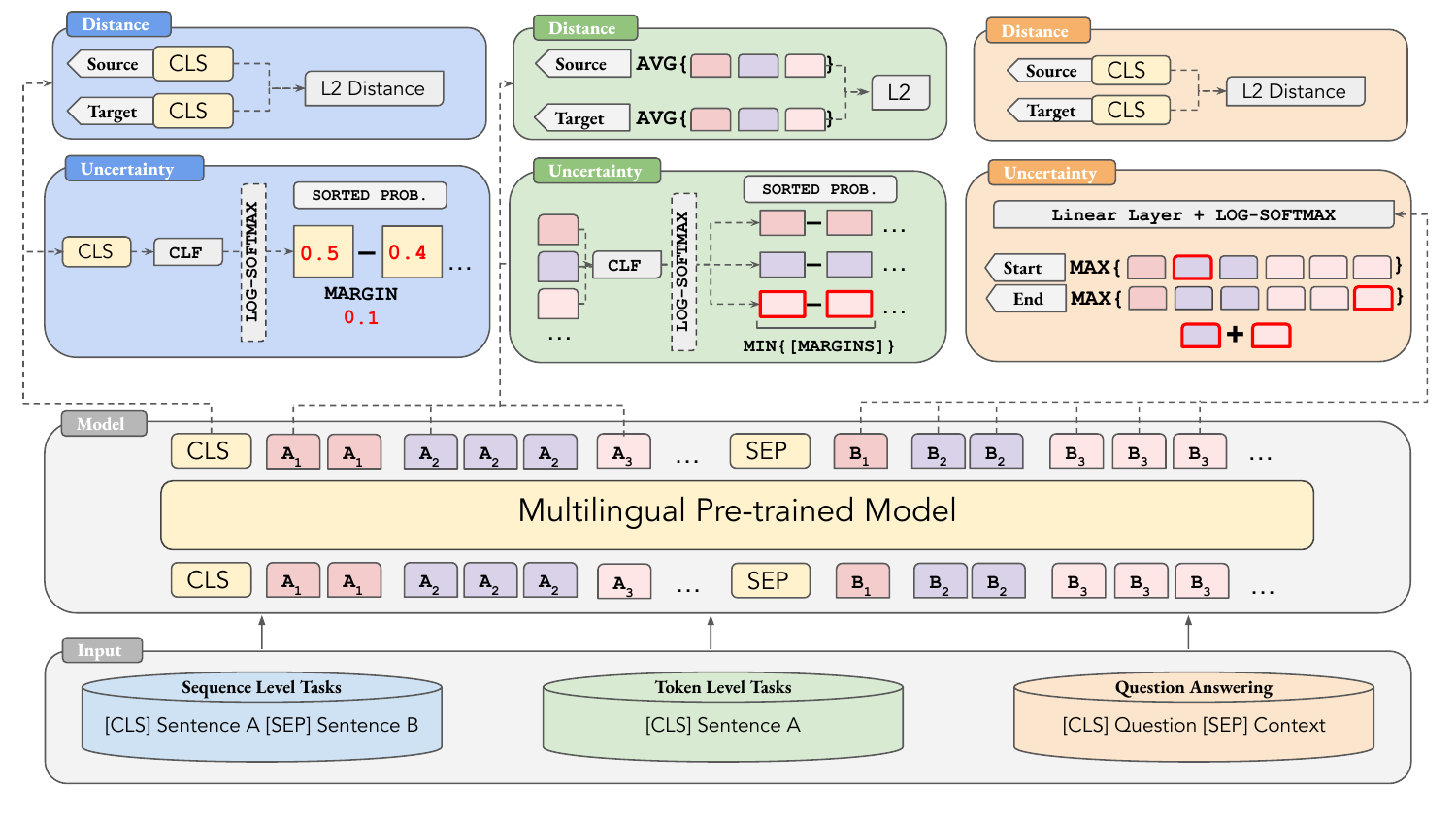}
    \caption{An overview of how distance and uncertainty are measured in our setup. $A_{1}, A_{2}, A_{3}$ denote three words in \emph{Sentence A} that are tokenized into 2, 3, and 1 subword, respectively.}
    \label{fig:overview}
\end{figure*}

\subsection{Distance and Uncertainty Measurements}
\label{app:dist_unc}
All of our strategies our based on reliable distance and uncertainty measures. Once these are established, it is easy to extend \method{} to other tasks and models. An overview of how these are measured for the tasks in our study can be found in \reffig{fig:overview}. These are formally described in the paper, in Section \ref{sec:strategies}.

\subsection{Uncertainty Details}
For token-level tasks,  we investigate two strategies. First, we employ the Mean Normalized Log Probability (\texttt{MNLP}) \cite{shen2017deep} method, which has been demonstrated as an effective uncertainty measure for \emph{Named Entity Recognition} (NER). This approach selects instances for which the log probability of model prediction, normalized by sequence length, is the lowest. Formally,
\vspace{-2mm}
\begin{gather*}
     \mathbf{X}^{*} = \mathrm{argmin}_{~\mathbf{X} ~\in~ \mathcal{S}^{b}} ~ \sum_{x_{s} \in \mathbf{X}}\texttt{MNLP}\left(x_{s}\right) \\
    \text{Where}\\
    \texttt{MNLP}\left(x\right) = \frac{1}{\lvert x \rvert} \sum_{i = 1}^{\lvert x \rvert} \log p^{i}_{c_1}(x)
\end{gather*}

In this equation, $\lvert x \rvert$ denotes the sequence length of the unlabeled sample $x$, and $p^{i}_{c_1}(x)$ represents the predicted probability of the most probable class for the $i^{th}$ token in the sequence.

Concurrently, we also explore margin-based uncertainty techniques (\texttt{MARGIN-MIN}). For each token in the sequence, we compute the margin as the difference between the probabilities of the top two classes. Then, we assign the minimum margin across all tokens as the sequence margin score and choose sequences with the smallest margin score. Formally,
\vspace{-5mm}

\begin{gather*}
     \mathbf{X}^{*} = \mathrm{argmin}_{~\mathbf{X} ~\in~ \mathcal{S}^{b}} ~ \sum_{x_{s} \in \mathbf{X}}\texttt{MARGIN-MIN}\left(x_{s}\right) 
    \\
    \text{Where}\\
    \texttt{MARGIN-MIN}\left(x\right) = \min_{i = 1}^{\lvert x \rvert} \left( p^{i}_{c_1}(x) - p^{i}_{c_2}(x) \right)
\end{gather*}
We eventually choose the margin based technique given better performance for both token level tasks.

\begin{table*}[ht]
\small
\centering
\begin{tabular}{c | c | c | c | c | C{6cm}}
\toprule
\multirow{2}{*}{\textbf{Dataset}} & \multicolumn{3}{c}{\textbf{Single Target}} & \multicolumn{2}{c}{\textbf{Multi-Target}} \\
\cmidrule(lr){2-4} \cmidrule(lr){5-6}
& \textbf{HP} & \textbf{MP} & \textbf{LP} & \textbf{Geo} & \textbf{LP Pool} \\ \midrule
UDPOS & fr:4938 & tr:984 & ur:545 & te:130, mr:46, ur:545 & ar:906, he:484, ja:511, ko:3014, zh:2689, fa:590, ta:80, vi:794, ur:545 \\ \midrule
NER & fr:9300 & tr:9497 & ur:944 & id:7525, my:100, vi:8577 & ar:9319,id:7525, my:100, he:9538, ja:9641, kk:910, ms:761, ta:965, te:939, th:9293, yo:93, zh:9406, ur:944 \\ \midrule
XNLI & fr:2490 & tr:2490 & ur:2490 & bg:2490, el:2490, tr:2490 & ar:2490, th:2490 ,sw:2490, ur:2490, hi:2490 \\ \midrule
TyDiQA & fi:1371 & ar:2961 & bn:478 & bn:478, te:1113 & sw:551, bn:478, ko:325 \\
\bottomrule
\end{tabular}
\caption{Number of unlabelled target examples used in each configuration. This is the size of the validation set.}
\label{tab:lang_number}
\end{table*}

\begin{table}
\small
\centering
\begin{tabular}{c | c | c | c }
\toprule
\textbf{Model} & \textbf{Dataset} & \textbf{LR} & \textbf{Epochs}  \\ \midrule
\multirow{4}{*}{XLM-R} & NER & 2e-5 & 10 \\
& UDPOS & 2e-5 & 10 \\
& XNLI & 5e-6 & 10 \\
& TyDiQA & 1e-5 & 3 \\ \cmidrule{1-4}
\multirow{4}{*}{InfoXLM} & NER & 2e-5 & 10 \\
& UDPOS & 2e-5 & 10 \\
& XNLI & 5e-6 & 10 \\
& TyDiQA & 1e-5 & 3 \\ \cmidrule{1-4}
\multirow{4}{*}{RemBERT} & NER & 8e-6 & 10 \\
& UDPOS &  8e-6 & 10 \\
& XNLI &  8e-6 & 10 \\
& TyDiQA & 1e-5 & 3 \\
\bottomrule
\end{tabular}
\caption{Hyperparameter Details.}
\label{tab:hparams}
\end{table}

\begin{table*}[ht]
\centering
\small
\begin{tabular}{c | C{2cm} | C{2cm} | C{2cm} | C{3cm} | C{3cm}}
\toprule
\multirow{2}{*}{\textbf{Dataset}} & \multicolumn{3}{c}{\textbf{Single Target}} & \multicolumn{2}{c}{\textbf{Multi-Target}} \\
\cmidrule(lr){2-4} \cmidrule(lr){5-6}
& \textbf{HP} & \textbf{MP} & \textbf{LP} & \textbf{Geo} & \textbf{LP Pool} \\ \midrule
\multirow{2}{*}{UDPOS} & fr & tr & ur & te, mr, ur & ar, he, ja, ko, zh, fa, ta, vi, ur \\
& it: 3434 & et:799, fa:598, ja:890, vi:694 & de:5095, eu:84, hi:3323, mr:92, nl:556 & ar:93, bg:68, et:3197,eu:84, fr:65, hi:3323, ja:890 & hi:830 \\ \midrule
\multirow{2}{*}{NER} & fr & tr & ur & id, my, vi & ar, id, my, he, ja, kk, ms, ta, te, th, yo, zh, ur \\
& it: 572 & he: 4582; mr: 1015, nl: 4393 & fa: 4007, ms: 79, ta: 788, vi: 62 & bn:1000, he:572, ko:469, ms:79 & bn:62, ko:4699, ml:4306 \\ \midrule
\multirow{2}{*}{XNLI} & fr & tr & ur & bg, el, tr & ar, th, sw, ur, hi \\
& ur:781, zh:6250 & ru:781, ur:6250 & hi:6250 & ru:781, ur:6250 & bg:6250, de:781, fr:781, vi:781 \\ \midrule
\multirow{2}{*}{TyDiQA} & fi & ar & bn & bn, te & sw, bn, ko \\
& ar:185, bn:119, id:142, ko:2, ru:1, sw:17, te:4450 & bn:14, fi:2742, ko:20, sw:4, te:2 & ar:740 & ar:46, fi:2742, id:570, ko:325, ru:5, sw:8 & - \\
\bottomrule
\end{tabular}
\caption{LITMUS prescribed annotation budget: LITMUS prescribes how many samples to select from each language. We select a random sample of data following the prescribed annotation.}
\label{tab:litmus}
\end{table*}

\subsection{Fine-tuning Details}
\label{app:ft}
As mentioned in the text, we first fine-tune each model on EN using the hyperparameters mentioned in the paper. Hyperparameters are in Table \ref{tab:hparams} and we report average results across three seeds: 2, 22, 42. For UDPOS, we include all languages except Tagalog, Thai, Yoruba and Kazakh, because they do not have training data for the task\footnote{\href{https://huggingface.co/datasets/xtreme/blob/main/xtreme.py\#L914}{https://huggingface.co/datasets/xtreme/blob/main/xtreme.py\newline\#L914}}.

% The detailed results for all strategies can be found in Tables \ref{tab:udpos_all}, \ref{tab:panx_all}, \ref{tab:xnli_all} and \ref{tab:tydi_all}. 
\subsection{Detailed Results}
\label{app:detailed}
The detailed results for the first ablation study where we test \method{} for multiple budgets in one AL round, can be found in Table \ref{tab:mul-bud}. Results for the second ablation, where we fine-tune models on randomly selected data that follows the same language distribution as \method{}, can be found in Table \ref{tab:same-ratio}.

\begin{table}
\small
\begin{tabular}{@{}c|c|ccccc@{}}
\toprule
\multirow{2}{*}{\textbf{Dataset}} & \multirow{2}{*}{\textbf{Strategy}} & \multicolumn{5}{c}{\textbf{AL Round}}              \\  \cmidrule{3-7}
                         &                           & 1      & 2      & 3      & 4      & 5      \\ \midrule
\multirow{3}{*}{PAN-X}   & SR                & 81.1 & 81.9 & 84.1 & 85.1 & 84.0 \\
                         & \method{}                     & 83.2 & 83.1 & 84.1 & 85.8 & 85.2 \\ \cmidrule{2-7}
                         & $\Delta$                    & \textbf{2.0}  & \textbf{1.2}  & 0.0  & \textbf{0.7}  & \textbf{1.2}  \\  \cmidrule{1-7}
\multirow{3}{*}{UDPOS}   & SR                & 89.9 & 89.3 & 89.5 & 90.0 & 89.8 \\
                         & \method{}                     & 89.7 & 89.8 & 89.9 & 89.5 & 90.1 \\ \cmidrule{2-7}
                         & $\Delta$                    & -0.2 & \textbf{0.5}  & \textbf{0.5}  & -0.4 & \textbf{0.3}  \\ \cmidrule{1-7}
\multirow{3}{*}{XNLI}    & SR                & 73.3 & 73.8 & 73.8 & 73.8 & 73.9 \\
                         & \method{}                      & 74.5 & 74.7 & 75.5 & 75.3 & 75.3 \\ \cmidrule{2-7}
                         & $\Delta$                     & \textbf{1.2}  & \textbf{0.9}  & \textbf{1.6}  & \textbf{1.5}  & \textbf{1.4}  \\ \cmidrule{1-7}
\multirow{3}{*}{TyDiQA}  & SR                & 80.6   & 81.5   & 81.5   & 82.0   & 81.7   \\
                         & \method{}                      & 82.8   & 83.2   & 83.2   & 83.5   & 83.8   \\ \cmidrule{2-7}
                         & $\Delta$                     & \textbf{2.2}    & \textbf{1.7}    & \textbf{1.8}    & \textbf{1.5}    & \textbf{2.1}   \\ \bottomrule
\end{tabular}
\caption{Detailed results: SR stands for SAME-RATIO. Same data distribution across languages but a random subset of datapoints selected.}
\label{tab:same-ratio}
\end{table}

% \usepackage{booktabs}
% \usepackage{multirow}
\begin{table*}
\centering
\small
\begin{tabular}{@{}l|r|c|c|c|c|c@{}}
\toprule 
 \multicolumn{1}{c}{\multirow{2}{*}{\textbf{Dataset}}} &
  \multicolumn{1}{c}{\multirow{2}{*}{\textbf{Budget}}} &
  \multicolumn{5}{c}{\textbf{Strategy}} \\ \cmidrule{3-7}
\multirow{8}{*}{UDPOS} &
  \multicolumn{1}{c}{} &
  \multicolumn{1}{l}{\texttt{GOLD}} &
  \multicolumn{1}{l}{\texttt{EGAL}} &
  \multicolumn{1}{l}{\texttt{KNN-UNC}} &
  \multicolumn{1}{l}{\texttt{AVG-DIST}} &
  \multicolumn{1}{l}{\texttt{UNC}} \\ \midrule
                        & 5      & 77.1& 70.9& 79.8& 81.8& 65.7\\
                        & 10     & 80.1& 70.7& 81.8& 82.3& 65.4\\
                        & 50     & 83.5& 72.4& 83.2& 83.8& 71.1\\
                        & 100    & 86.5& 74.5& 86.0& 85.8& 75.1\\
                        & 250    & 89.6& 77.7& 88.2& 88.2& 77.4\\
                        & 500    & 91.1& 79.1& 89.3& 89.3& 79.7\\
                        & 1000   & 92.2& 81.5& 89.3& 89.5& 82.2\\ \cmidrule{1-7}
\multirow{7}{*}{PAN-X}  & 5      & 64.3& 54& 64.9& 40.2& 40.8\\
                        & 10     & 71.8& 52.7& 58.4& 44.2& 47.9\\
                        & 50     & 77.1& 61.6& 74.5& 65.4& 46.8\\
                        & 100    & 79.2& 67.6& 80.5& 70.2& 37\\
                        & 250    & 82.9& 76.7& 82.4& 76.7& 61.1\\
                        & 500    & 85.7& 80.1& 84.5& 82.5& 73.6\\
                        & 1000   & 87.7& 80.6& 83.5& 81.7& 79.1\\ \cmidrule{1-7}
\multirow{7}{*}{XNLI}   & 5      & 64.3& 63.6& 69& 57.1& 67.6\\ 
                        & 10     & 70.6& 68.1& 70.5& 70.8& 72.5\\
                        & 50     & 73.1& 73.3& 73.2& 72.6& 72.2\\
                        & 100    & 72.9& 71.2& 72.7& 71.1& 73.8\\
                        & 250    & 71.5& 72.4& 72.4& 72.9& 73.9\\
                        & 500    & 73.3& 72.8& 72.3& 72.4& 72.6\\
                        & 1000   & 73.9& 72.3& 72.3& 73.7& 72.3\\ \cmidrule{1-7}
\multirow{7}{*}{TyDiQA} & 74.5& 73.3& 73.4& 72.9& 74.2& 74.2   \\
                        & 74.1& 73.3& 74.0& 73.8& 74.4& 74.4   \\
                        & 75.5& 75.5& 77.1& 73.0& 75.3& 75.3   \\
                        & 77.6& 74.6& 80.2& 73.7& 77.8& 77.8   \\
                        & 80.6& 74.9& 80.6& 76.8& 78.8& 78.8   \\
                        & 82.2& 78.1& 81.1& 78.7& 81.6& 81.6   \\
                        & 82.8& 79.9& 80.4& 80.8& 81.5& 81.5   \\ \bottomrule 
\end{tabular}
\caption{Detailed results: Multiple budgets, one AL round}
\label{tab:mul-bud}
\end{table*}

\end{document}